\documentclass[letterpaper, 10 pt, conference]{ieeeconf}  

\IEEEoverridecommandlockouts    

\usepackage{graphicx}
\usepackage{booktabs}
\usepackage[english]{babel}
\usepackage{multirow}
\usepackage{tabularx}
\usepackage{pgfplots}
	\pgfplotsset{compat=newest}
\usepackage{mathrsfs}
\usepackage{cite}
\usepackage{balance}
\usepackage{amsmath}
\usepackage{amsfonts}

\usepackage[font=small,skip=4pt]{caption} 

\begin{document}

\title{\LARGE \bf Compact Deep Neural Networks for Computationally Efficient Gesture Classification From Electromyography Signals}

\author{Adam~Hartwell,~Visakan~Kadirkamanathan,~and~Sean~R.~Anderson
\thanks{A. Hartwell, V. Kadirkamanathan and S. R. Anderson are with the Department of Automatic Control and Systems Engineering, University of Sheffield, Sheffield, S1 3JD
UK, e-mail: ahartwell1@sheffield.ac.uk}
}

\maketitle
\thispagestyle{empty}
\pagestyle{empty}

\begin{abstract}
Machine learning classifiers using surface electromyography are important for human-machine interfacing and device control. Conventional classifiers such as support vector machines (SVMs) use manually extracted features based on e.g. wavelets. These features tend to be fixed and non-person specific, which is a key limitation due to high person-to-person variability of myography signals. Deep neural networks, by contrast, can automatically extract person specific features - an important advantage. However, deep neural networks typically have the drawback of large numbers of parameters, requiring large training data sets and powerful hardware not suited to embedded systems. This paper solves these problems by introducing a compact deep neural network architecture that is much smaller than existing counterparts.  The performance of the compact deep net is benchmarked against an SVM and compared to other contemporary architectures across 10 human subjects, comparing Myo and Delsys Trigno electrode sets. The accuracy of the compact deep net was found to be $84.2\pm6\%$ versus $70.5\pm7\%$ for the SVM on the Myo, and $80.3\pm7\%$ versus $67.8\pm9\%$ for the Delsys system, demonstrating the superior effectiveness of the proposed compact network, which had just 5,889 parameters - orders of magnitude less than some contemporary alternatives in this domain while maintaining better performance. 
\end{abstract}
\IEEEpeerreviewmaketitle

\section{Introduction} \label{Intro}

    Machine learning is an essential tool for extracting user intention from bio-electric signals for control of devices \cite{AsghariOskoei2007, farina2014extraction}. In the domain of hand movements, classification from surface electromyography (sEMG) has been performed using methods such as support vector machines (SVMs) \cite{Atzori2014d, Castellini2009, Lucas2008, quitadamo2017support}, neural networks \cite{Atzori2014d, Castellini2009, Duan2016}, neurofuzzy \cite{Balbinot2013, Khezri2011} and mixtures of experts \cite{baldacchino2018simultaneous}. Typically, for these conventional classifiers, feature extraction is performed using manually chosen features e.g. wavelets or Fourier transforms \cite{Canal2010, Reaz2006}. This approach to feature extraction is limited because the features are not person-specific. This is important because there is a large variability of myoelectric activity from person-to-person, suggesting that features should be tuned specifically to each individual.
    
    Deep neural networks \cite{LeCun2015, Krizhevsky2012} are potentially advantageous for gesture recognition because they can perform person-specific feature extraction. The networks can be trained from raw sEMG data which allows learning of features tailored to each subject during feature extraction in the early layers which contrasts hand designed features which are generally computed in the same manner on all subjects.
    
    There are now a few instances of deep neural networks used in hand movement classification from sEMG \cite{Atzori2016, geng2016gesture, Pizzolato2017, zhai2017self}. However, one of the barriers to the wider uptake of deep neural networks is the typical large network size and associated large number of parameters. This makes it difficult to ensure good generalisation of the trained network, and also sets out a requirement for large training data sets, which can be difficult and time-consuming to obtain. In addition, large networks are not well suited to real-time implementation in embedded systems using portable GPUs with relatively few cores (e.g. 256 CUDA cores on the NVIDIA Jetson TX2 for embedded systems, compared to 3,500 CUDA cores on the NVIDIA GeForce GTX 1080 Ti desktop GPU). 
    
    In this paper, we propose the use of compact deep convolutional neural networks (CNNs) for gesture recognition, in order to solve the problems associated with large deep networks. Recently, model compression techniques have been developed for deep nets to massively reduce their size, using singular value decomposition \cite{denton2014exploiting}, network pruning \cite{han2015learning} and deep compression \cite{han2015deep}. One method, SqueezeNet, has produced state-of-the-art results compared to those techniques in image processing \cite{Iandola2016}. We modify the SqueezeNet architecture here and encode a spatial-reduction strategy into the network to produce a novel CNN architecture that has far fewer parameters than those used previously in this domain (Fig. 1). This solves a number of problems associated with large network size, thus opening up wider potential for person-specific machine learning algorithms for recognition problems.
    
    To evaluate the compact CNN, we benchmarked against an SVM with wavelet features on a hand movement classification task using sEMG data. The experimental data was obtained from 10 human subjects. We also compared high grade sEMG electrodes from Delsys against the low cost Myo armband (Fig. 2). The number of hand movement gestures classified was 15 (Fig. 2). The results demonstrate that the compact deep net outperforms the SVM for all human subjects, for each electrode type.
    
    Further we compared our compact deep neural network to other contemporary deep neural network architectures in terms of accuracy, number of parameters and run-time performance on both an NVIDIA Jetson TX2 GPU and an NVIDIA GeForce GTX 1080 Ti. Our results demonstrate that we maintain high performance (better than contemporary, large, deep neural network architectures) whilst significantly reducing run-times on embedded hardware.

\section{Methodology} \label{Methods}
  
  In this section we describe the compact deep neural network, the SVM, the performance evaluation (including cross validation approach), experimental data collection and our runtime evaluation strategy.
  
    \begin{figure}
        \vspace{0.7em}
        \centering
        \includegraphics[width=\columnwidth]{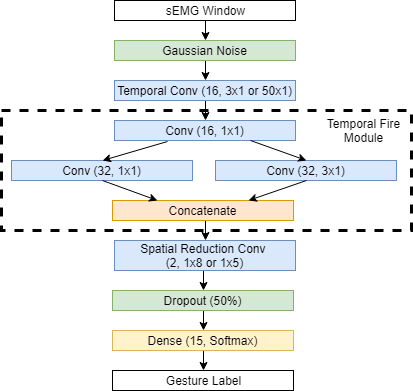}
        \caption{Neural network architecture for Myo and Delsys data respectively. Note that the Temporal Convolution stage has $3\times1$ size filters for the Myo electrodes and $50\times1$ for the Delsys Trigno electrodes (due to the higher sample frequency of the Delsys system). The Spatial Reduction Convolution is 1x8 for the Myo data and 1x5 for the Delsys data due to the number of input channels.}
        \label{fig:net_diagram}
    \end{figure}

    \subsection{Deep Convolutional Neural Network} \label{sec:net-design}
        The input to the CNN is a window of sEMG data, $X\in\mathbb{R}^{n_s\times n_c}$, where $n_s$ is the number of samples and $n_c$ the number of sEMG channels. The main building block of the CNN is the convolutional layer, where a 2D convolution is a single 2D map, indexed by $k$, in layer $l$, is $Z^{(l,k)} \in \mathbb{R}^{r_l \times c_l}$, where $Z^{0,1}=X$. At each layer there is a stack of $d_l$ maps, i.e. a 3D volume of dimension $r_l\times c_l \times d_l$. The value of a unit, $z_{r,c}^{(l,k)}$, at location $(r,c)$, in the map $Z^{(l,k)}$, is given by
        \begin{equation} 
             z_{r,c}^{(l,k)} = h_a \left( \left( \sum_{m=1}^{d_{l-1}}  \sum_{i=1}^{R_l} \sum_{j=1}^{C_l} w_{i,j}^{(l,k,m)} z^{(l-1,m)}_{\tilde{r} + i, \tilde{c} + j} \right) + b^{(l,k)}   \right) 
        \end{equation}
        where $z_{r,c}^{(l,k)}$ is the neuron output at location $(r,c)$, for $r=1,\ldots,r_l$, $c=1,\ldots,c_l$, $R_l \times C_l$ is the convolution filter size, the convolution filter indexed by $k$, for $k=1,\ldots,d_l$, is composed of the adjustable CNN weights $w_{i,j}^{(l,k,m)}$, $b^{(l,k)}$ is a bias term, and $\tilde{r}=r - \lceil R_l/2 \rceil$ and $\tilde{c}=c - \lceil C_l/2 \rceil$ for odd valued $R_l$ and $C_l$. $h_{a}(.)$ is the activation function of the neuron, defined here, for all but the final layer, as the leaky rectified linear unit (LReLU) \cite{maas2013rectifier, Xu2015}, where
        \begin{equation} 
            h_{a}(x) = 
            \begin{cases}
                x, \ \ \ x \geq 0\\
                \alpha x,\ x < 0\\
            \end{cases}
        \end{equation}
        where $0 < \alpha < 1$ but is generally a small value such as $0.1$ which is used here.
        
        The final layer, which performs the classification, is a dense layer with softmax activation function, 
        \begin{equation} 
             z^{*}_j = \exp \left( \tilde{z}_j^{(l)} \right) \times \left( \sum^{M - 1}_{k = 0} \exp \left( \tilde{z}_k^{(l)} \right) \right)^{-1} 
        \end{equation}
        for class $j=1,\ldots,M-1$, where $M$ is the number of classes,  $z^{*}_j$ is the normalised output of the softmax layer for class $j$ and
        \begin{equation}
            \tilde{z}_r^{(l)} = \left( \sum_{m=1}^{d_{l-1}} \sum_{j=1}^{r_{l-1}} \sum_{i=1}^{c_{l-1}} w_{i,j}^{(l,m)} z^{(l-1,m)}_{i,j} \right) + b^{(l,r)}
        \end{equation}
        for $r=1,\ldots,M-1$.
        
        The network weights were trained using the following cross-entropy loss function, for $N$ data samples and $M$ classes,
         \begin{equation} 
            L(\Theta) = - \sum_{i=1}^{N} \sum_{j=0}^{M-1} 1 \lbrace y^{(i)} = j \rbrace \log z^{*}_{ij} 
        \end{equation}
        where $\Theta$ is the set of all CNN parameters, including weights and biases from all layers, $z^{*}_{ij}$ is the softmax output for data sample $i$ and prediction of class $j$, $y^{(i)}$ is the true class label for data sample $i$,  $1\{.\}$ is the indicator function, i.e. $1\{.\}=1$ for true and $1\{.\}=0$ for false. The cross-entropy was minimised using the Adam algorithm, which is a variant of first order stochastic gradient descent with momentum \cite{Kingma2015}. The weight parameters were randomly initialised using the Glorot uniform kernel \cite{Glorot2010}, and bias parameters were initialised to zero. To prevent overfitting, dropout regularisation was used (Fig. 1) and early stopping on validation data via a minimum improvement threshold of 0.5\% for 5 iterations.
        
        A key novel feature here is the architecture based on SqueezeNet. The ``Temporal Fire Module'' indicated in Figure \ref{fig:net_diagram} is a customisation of the ``Fire Module'' used in the SqueezeNet network design \cite{Iandola2016}, which allows high performance classification while keeping the number of parameters low. Our variant expands only in the temporal direction of the sEMG data enforcing the extraction of low-level temporal features, whilst maintaining a high performance-to-parameters ratio. This design was found to produce results significantly better than a less constrained approach which allows early features to be spatial-temporal or only spatial in nature.
        
        The other key novel architecture choice is the spatial reduction convolution placed late in the network (Fig. 1). This spatial convolution uses only $2$ filters thus explicitly encoding that we expect there to be few spatial combinations that are meaningful for determining the gesture classes. This allows a significant reduction in the number of parameters but again with only minimal performance loss.
        
        The intuition behind the spatial reduction is that different sensors will be more important on different subjects due to physical and biological differences as well as issues such as cross-talk and so this layer acts a filter selecting which channels are most important to the classification task.
        
        We implemented the CNN in Keras \cite{chollet2015keras}, which is a Python front-end for designing deep neural networks, which here was used with the Python library Tensorflow \cite{tensorflow2015-whitepaper} for computational implementation. Parameter estimation (training) for the CNNs was performed using an NVIDIA Tesla K40 GPU with 12 GB RAM. Note that although a high performance GPU was used here for rapid training, implementation was evaluated on a NVIDIA Jetson TX2 GPU (256 CUDA cores) designed for embedded systems and an NVIDIA GeForce GTX 1080 Ti (3,500 CUDA cores).

\subsection{Support Vector Machine}

    The SVM was designed using a Radial Basis Function kernel, a one-vs-all approach was used to hand the multiple classes, the gamma factor was set to the reciprocal of the number of features and class weighting was inversely proportional to number of examples, no probability estimates were used. The marginal Discrete Wavelet Transform (mDWT) \cite{Lucas2008} down to the 3rd level of decomposition was used as the feature representation due to its previously good performance in similar studies \cite{Atzori2014d}.

\subsection{Generic CNN Benchmark}
    A generic CNN with no domain-based optimisations was also evaluated to give a baseline in terms of macro accuracy potential and run-time performance. This CNN represented a naive, large, deep neural network solution without the compactness offered by the SqueezeNet architecture. 
    
    This CNN was structured as 12 convolutional blocks: each block consisted of a 1x1 or 3x3 convolution with 32 filters and a stride of 1, followed by batch normalisation and finished with a Leaky ReLU with $\alpha=0.1$. The block order was 3x3, 3x3, 3x3, 1x1 and this order was then repeated twice more creating an architecture made of 12 blocks. After the 12th block a dense layer was connected using a softmax activation to generate the final classification output. The same early stopping criteria and initialisation scheme was used as described in Section \ref{sec:net-design} to guard against overfitting.

    \subsection{Experimental Data} 
        \subsubsection{Overview}
        
            All data were collected under approval of the University of Sheffield ethics board. Our study gathered data from the dominant hand of 10 healthy subjects using both a Myo Armband (8 surface electrodes) \cite{ThalmicLabs2015} and 5 Delsys Trigno \cite{Delsys2017} wireless surface electrodes. The Myo and Delsys electrodes were worn simultaneously to record exactly the same movement signals. This had the consequence that the electrode sets could not be placed in the same spatial location. However, we judged that the advantage of recording the same movement out-weighed this drawback. In addition, the Myo armband is  limited in its placement, whilst the Delsys electrodes are much more flexible and can arguably be more effectively targeted to specific muscle groups useful for hand movement classification. 
            
            Each subject performed 6 repetitions of 14 gestures (+\textit{rest}) holding each for 10 seconds. A strong timing delimitation was used to ensure all data labelled for a gesture only contained the gesture itself and not the movement into or out of the gesture. This was found to produce a much more reliable online classification result than attempting to train a classifier that attempted to learn these edge effects. 
        
        \subsubsection{Electrode placement}
        
            The Myo Armband was placed $2/3$s of the way up the forearm (measured from lower electrode edge) with main electrode block directly on top, status LED closest to the wrist, band perpendicular to forearm as this is the placement recommended by the manufacturer.
            
            The Delsys Trigno electrodes were placed using the sticky patches recommended by the manufacturer. These positions were selected to target both specific muscles and general areas of interest. Electrode E1 was placed just behind the wrist along the Abductor Pollicis Longus muscle. E2 was placed similarly behind the wrist along the Flexor Digitorum Superficialis. E3  was set behind the wrist along the Extensor Carpi Ulnaris. E4 was placed further up the forearm, but in front of the Myo Armband along the Flexor Carpi Radialis. Lastly, electrode E5 was placed in-line with E4 up the forearm but along the Flexor Carpi Ulnaris (Fig. 2).  
            
            Timestamps were used to synchronise data streams for labelling and a sliding window of length $\sim150$ms (due to changing sampling rates) with an increment of $\sim5$ms was used. An exception is made for re-implementation of the Geng et al. \cite{geng2016gesture} network as it performs instantaneous classification therefore requiring a window length of 1 sample.
        
            \begin{figure}
                \vspace{0.7em}
                \centering
                \includegraphics[width=\columnwidth]{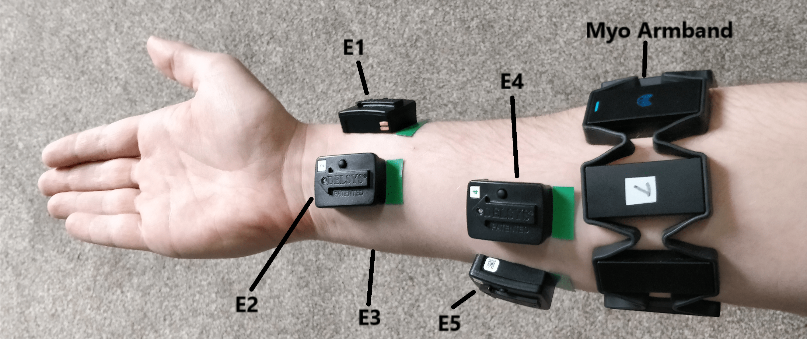}
                \includegraphics[width=\columnwidth]{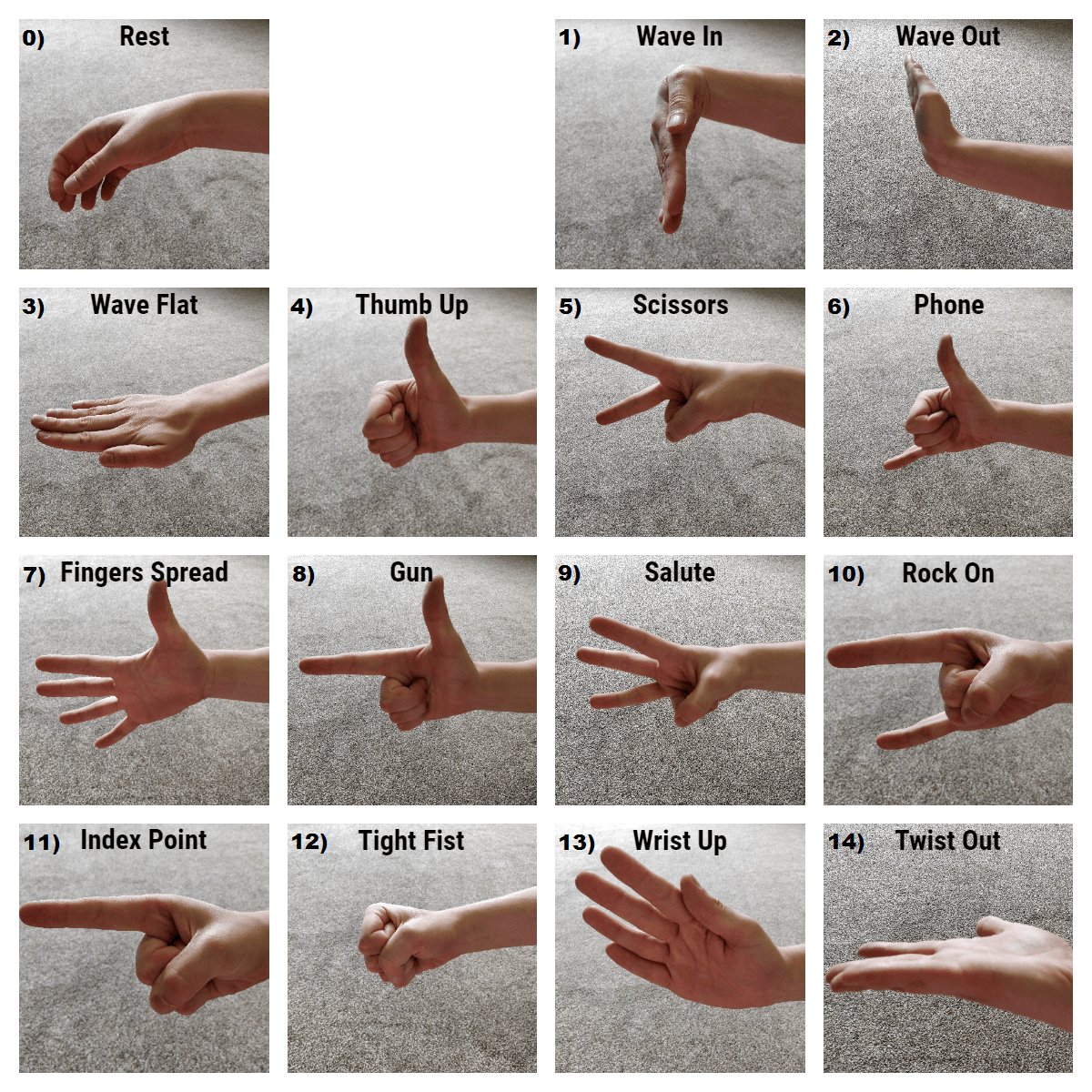}
                \caption{Position of electrodes on the forearm of subject 1 (top) and the 14 gestures (+ \textit{rest}) included in the study and their associated labels (bottom).}
                \label{fig:electrodes_gestures}
            \end{figure}

        \subsubsection{Gesture set selection}
        
            The set of gestures used were selected from a large pool of candidates based on the hand taxonomy literature, recognisable gestures (such as from sign languages) and commercial sEMG work. Selection criteria was based on preliminary trials conducted on subject 1. These trials involved gathering data on each of these different gestures and quantitatively comparing the performance achieved by the Neural Network and SVM classifiers on offline data with various combinations of these gestures as well as a qualitative comparison of the classification potential in an online context using the Myo Armband. 

        \subsubsection{Gesture movement}    
            A stationary hold of a gesture for 10 seconds was chosen over a more rapid movement into and out of a gesture to avoid the issue of inaccurate ground truth caused by movement. When a subject performs a movement as opposed to a hold the classification algorithm must account for the movement into, hold of the gesture and movement out of the gesture back to \textit{rest} (typically) which is highly likely to violate the explicit assumption in most classification algorithms that each class is unique in some way because the movements into and out of each gesture are likely to be similar.

\subsection{Performance Evaluation}
    We use the macro average accuracy as our keystone performance metric \cite{Sokolova2009}, which weights all classes equally,
    \begin{equation} 
        \bar{a}_{ma} = \frac{1}{G} \sum^{G}_{i=1} \frac{\text{TP}_{i}}{\text{TP}_{i} + \text{FN}_{i}}
        \label{macro_acc}
    \end{equation}
    where $\bar{a}_{ma}$ is the macro-average accuracy, FN$_{i}$ is the false negatives of gesture $i$, TP$_{i}$ is the true positives for gesture $i$ and $G$ is the number of gestures (classes).  
    
    This metric encodes explicitly the idea that all classes are equally important and helps provide an unbiased measure of performance, it would also help account for any imbalance in the number of training examples although this is not an issue here due to our experimental methodology.

    We perform a variation of 12-fold stratified cross validation on each subject to acquire a mean performance for each subject using totaled $TP$ and $FN$ for a less biased result \cite{Forman2010} as in Equation \ref{cross-val}. We then calculate the mean of this across all subjects to get an estimate for the expected performance on a new subject. 
    
    \begin{equation} 
        \bar{a}_{ma}^* = \frac{1}{G} \sum_{i=1}^G \frac{  \sum_{j=1}^K \text{TP}_{i,j}}{ \sum_{j=1}^K \text{TP}_{i,j} + \sum_{j=1}^K \text{FN}_{i,j}} 
        \label{cross-val}
    \end{equation}
    where $K$ is the number of cross validation folds.
    
    Typically when selecting folds in cross validation samples are selected randomly however in this dataset and in any other data set where a sliding window is used with an increment lower than the window length this is inappropriate and undermines any conclusions drawn because random selection will include windows that overlap violating training-test data separation.
    
    In order to avoid this problem we split data via repetition number into training (3/6), validation (1/6) and test sets (2/6). This method ensures no informational contamination between between sets. 
    
    Numerically, for the Myo data this led to $\sim90,000$ training samples, $\sim30,000$ validation samples and $\sim60,000$ independent test samples. The Delsys data was roughly $\sim87,000$, $\sim29,000$, $\sim58,000$ due to the longer window length in terms of samples.
    
    Validation data was used for early stopping during neural network training and ignored for SVMs to ensure comparability between results. Each classifier was trained independently on each subject and the mean performance was then calculated across all subjects and validation folds.
 
\subsection{Run-time Performance}
    We used two different computing platforms to evaluate runtime performance in a real world context. One was a general purpose computer with an Intel Core i5-6500 and an NVIDIA GTX 1080 Ti GPU with 3,500 CUDA cores. The second was the NVIDIA Jetson TX2 - a low power embedded device designed for use with neural networks with just 256 CUDA cores. Both ran Ubuntu 16.10 LTS, Tensorflow 1.5.0 \cite{tensorflow2015-whitepaper} and Keras 2.0.6 \cite{chollet2015keras}.
    
    We made use of Python's ``timeit'' package reporting the lowest value from 20 trials, each of which took the mean run-time of 1000 predictions. This produced a soft lower bound on computation time. The upper bound, mean and standard deviation of trials were less informative here due to interactions with the operating systems and other running programs.

\section{Results and Discussion} \label{Results} \label{Discussion}

\begin{figure*}[ht!]
        \centering
        \includegraphics[width=\textwidth]{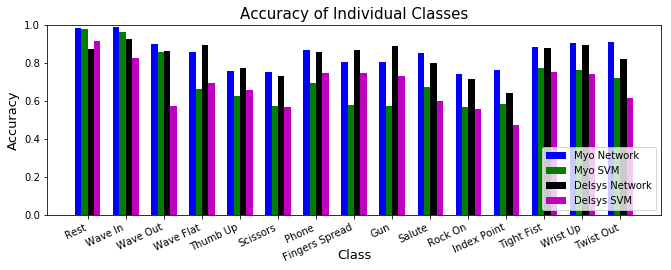}
        \caption{Comparison of accuracy per class of the neural networks and SVMs trained on the Delsys and Myo data.}
        \label{fig:myo_v_delsys_net}
        \vspace{-2.5ex}
    \end{figure*}

\begin{figure*}[ht!]
        \centering
        \includegraphics[scale=0.9]{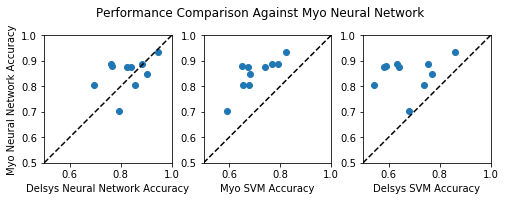}
        \caption{Per Subject performance comparison between compact CNN and SVM classification algorithms, for both Myo and Delsys data sets. Note that each dot represents an individual test subject.}
        \label{fig:XvY}
    \end{figure*}

    We compared the classification performance of the compact CNN to the SVM, which showed that the CNN outperformed the SVM for all movements (Figure \ref{fig:myo_v_delsys_net}), and all subjects (Figure \ref{fig:XvY}), on both Myo and Delsys data sets.  To summarise, the macro accuracy on the Myo data across all subjects and movements was $84.2\pm6\%$ for the deep net versus $70.5\pm7\%$ for the SVM. For the Delsys data the macro accuracy was $80.3\pm7\%$ using the deep net versus $67.8\pm9\%$ using the SVM. 
    
    We compared the size of this compact CNN to previously published types of deep neural network (Table I). The compact CNN used here contained only 5,889 parameters, which is far fewer than networks previously used in this domain.  Table I shows a comparison of the compact CNN against contemporary networks retrained and tested using our methodology and data. On a desktop PC using a GTX 1080 Ti we found, as expected, that number of parameters was not critical for fast run-time implementation. However, for the Jetson TX2, which is more representative for embedded systems, we observed significant improvements in run-time using the compact CNN.

    The generic CNN architecture actually performed best of all architectures on both data sets by $2-3\%$ but scaled poorly in number of parameters with the window length leading to a much slower run-time on the Delsys data (Table I). Our compact CNN performed second best overall, but still outperformed other published networks in terms of macro accuracy, and performed best in terms of run-time by a significant margin. This trade-off between high accuracy and fast run-time makes the novel, compact, deep neural network designed here the best suited to embedded systems. 
    
    
    \begin{table}
        \centering
        \begin{tabular}{|c|cccc|} 
        \hline
            \multicolumn{5}{|c|}{Myo Data} \\
        \hline
              & Params &  Acc.  & 1080 Ti & TX2 \\ 
        \hline
        Compact CNN   & 5,889 & 84.2\%  & 1.68ms & 7.89ms\\ 
        Atzori et al. \cite{Atzori2016}: Delsys  & 97,883 & 81.7\%  & 1.69ms & 13.17ms \\
        Generic CNN & 135,599  & 86.8\%  & 2.40ms  & 13.57ms \\
        Geng et al. \cite{geng2016gesture}     & 644,435 & 44.1\%  & 3.19ms & 22.26ms \\
        \hline
        \end{tabular}
        \\[2ex]
        \begin{tabular}{|c|cccc|} 
        \hline
            \multicolumn{5}{|c|}{Delsys Data} \\
        \hline
              & Params & Acc. & 1080 Ti  & TX2\\ 
        \hline
        Compact CNN   & 5,657 & 80.3\% & 1.74ms  & 8.07ms \\ 
        Atzori et al. \cite{Atzori2016}: Delsys  & 99,308 & 65.4\%  & 1.66ms & 15.36ms \\
        Generic CNN & 740,399  & 83.1\%  & 2.66ms & 24.55ms \\
        Geng et al. \cite{geng2016gesture}     & 546,131 & 26.4\%  & 3.21ms & 20.14ms\\
        \hline
        \end{tabular}
        \caption{Comparison of number of parameters, cross-subject mean macro accuracy and run-times for different types of deep convolutional neural network used in hand movement classification. }
        \vspace{-2.5ex}
    \end{table}
    
    The Geng et al. \cite{geng2016gesture} network performed with low accuracy here (Table I), which was  likely due to the network's use of a single sample of EMG as input, not a window, as was used in all other methods. This single sample approach appeared more successful in their previous work where there were large numbers of electrodes available \cite{geng2016gesture}. Also, our evaluation method took into account data balancing equally weighting performance across classes, which might not have been the case in their original study. 

    The results indicate that the Myo Armband can provide a level of performance similar to the Delsys Trigno system for the purpose of hand movement classification, albeit with more electrodes (Figure \ref{fig:XvY}). The Myo also has the advantages that it is easier to setup, easier to integrate with other applications and a factor of 100 cheaper. The Delsys system does provide a greater degree of flexibility in electrode placement and extensibility in the form of using more electrodes, however, which may allow a more complex setup to provide better performance than the Myo. 
    
    A similar study using deep nets with Myo and Delsys electrodes \cite{Pizzolato2017} indicated that the Delsys system improves, relative to the Myo, on classification performance for many movements ($\sim$50), likely due to its superior sampling and electrode quality. However, for situations where only 10-15 movements are required, the Myo is likely the better solution given it's comparable performance and lower cost.
    
\section{Conclusions} \label{Conclusions}
    We have presented a compact deep neural network based approach to gesture recognition. The compact CNN was evaluated on hand movement classification on inexpensive sEMG hardware in the form of the Myo Armband and compared to the Delsys Trigno electrodes. 
    
    We compared the compact deep neural network to an SVM-based approach using the features based on the marginal discrete wavelet transform and found our approach to be significantly better, achieving a $\sim15\%$ performance enhancement for both types of electrode.
    
    Lastly, we compared our compact architecture to other larger, contemporary deep neural network architectures demonstrating that our compact CNN performs better in terms of accuracy and run-time performance making it well suited to sEMG applications in general and especially to applications that have a restriction on computational power, such as embedded systems.

\bibliographystyle{IEEEtran} 
\bibliography{library.bib}

\end{document}